\documentclass[a4paper,conference]{IEEEtran}

\usepackage{cite}
\ifCLASSINFOpdf
\else
\fi
\usepackage{float}
\usepackage[utf8]{inputenc}
\usepackage{amsmath,amssymb,amsfonts, bbm}
\usepackage{graphicx,booktabs,array}
\usepackage{multirow}
\usepackage{makecell}
\usepackage{algorithm}
\usepackage{algpseudocode}
\usepackage{stfloats}

\begin{document}
%
\title{Transformers as Neural Augmentors: Class Conditional Sentence Generation via Variational Bayes}
\author{\IEEEauthorblockN{M. Şafak Bilici}
\IEEEauthorblockA{Department of Computer Engineering\\
Yildiz Technical University\\
Istanbul, Turkey\\
Email: safak.bilici@std.yildiz.edu.tr}
\and
\IEEEauthorblockN{Mehmet Fatih Amasyali}
\IEEEauthorblockA{Department of Computer Engineering\\
Yildiz Technical University\\
Istanbul, Turkey\\
Email: amasyali@yildiz.edu.tr}
}
\maketitle
\UseRawInputEncoding
\begin{abstract}
Data augmentation methods for Natural Language Processing tasks are explored in recent years, however they are limited and it is hard to capture the diversity on sentence level. Besides, it is not always possible to perform data augmentation on supervised tasks. To address those problems, we propose a neural data augmentation method, which is a combination of Conditional Variational Autoencoder and encoder-decoder Transformer model. While encoding and decoding the input sentence, our model captures the syntactic and semantic representation of the input language with its class condition. Following the developments in the past years on pre-trained language models, we train and evaluate our models on several benchmarks to strengthen the downstream tasks. We compare our method with 3 different augmentation techniques. The presented results show that, our model increases the performance of current models compared to other data augmentation techniques with a small amount of computation power.
\end{abstract}
\IEEEpeerreviewmaketitle
\section{Introduction}
In recent years, remarkable performance of Transformer \cite{NIPS2017_3f5ee243} based architectures are shown on downstream tasks with or without pre-training. They are widely used for downstream tasks as sentence classification, question answering, coreference resolution.  Besides that, using encoder-decoder Transformer model allows process sequence-to-sequence tasks like machine translation, summarization. In pre-training phase of these tasks, it is easy to collect a corpus since pre-training is done in self-supervised fashion, generally. However, at finetuning, it is required to collect significant number of samples with labels. One way to increase number of training samples is data augmentation. Nevertheless, data augmentation for NLP is not well-defined as in computer vision. If we treat token values as just a numbers in input, we reject the nature of the language. Hence, it is better to design more task-oriented data augmentation methods in NLP with linguistic validity.

To address the sentence classification problem, we introduce a novel generative data augmentation method for class conditional synthetic sentence generation. Our model is a modified version of original Transformer model, with a class conditional variational approximation between encoder and decoder. After training class conditional variational Transformer, synthetic sentences can be obtained with belonging desired class label. By doing this, distribution of generated sentences is not too irrelevant from the original sentences in desired class. 

\begin{figure*}[t]
    \centering
    \includegraphics[width=0.75\textwidth]{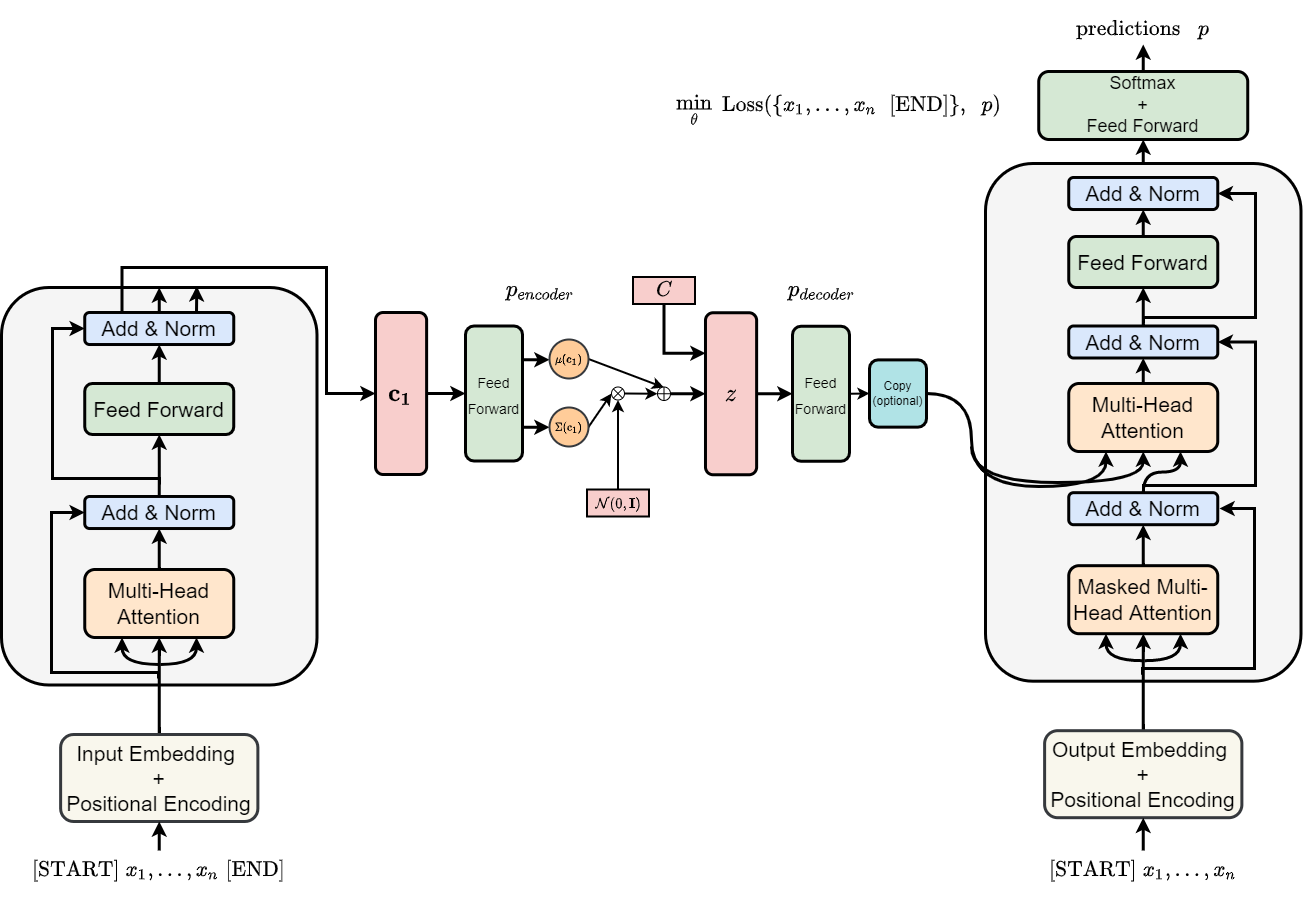}
    \caption{Model architecture of conditional variational Transformer.}
    \label{fig:cvaet}
\end{figure*}
Besides that, finetuning pre-trained models yet already computational expensive; hence, a data augmentation method should require low computational power. We show that our method works well with sub-sampling on used dataset.

We run our experiments on four datasets with different modifications and approaches on our model.
\section{Related Work}
\subsection{Data Augmentation for NLP}
In recent years, data augmentation is studied for natural language tasks. A traditional way to expand the existing data is using synonyms of words. This paraphrasing method can be done by a large lexical databases of languages, or by learning. As an example, in \cite{zhang2016characterlevel}, authors experimented data augmentation by using an English thesaurus. Also, there are several data augmentation methods based on word level \cite{xie2017data}, \cite{wei-zou-2019-eda}. Another way to use paraphrasing as data augmentation is back-translation \cite{sennrich2016improving} which can be used to obtain more diverse sentences as a neural data augmentation method. Following that, Transformers are used as generative data augmentation technique \cite{yang-etal-2020-generative}. In \cite{kafle-etal-2017-data}, authors proposed two task specific data augmentation methods for Visual Question Answering. Also, there are existing work on autoencoder based data augmentation. Authors of \cite{ng-etal-2020-ssmba} samples new sentences from a denoising autoencoder with masked language corruption. A latent space based generative approach is introduced in \cite{9599089}. They use variational autoencoder \cite{kingma2014autoencoding} with Gated Recurrent Unit \cite{cho2014learning}.
\subsection{Text Based Variational Autoencoder}
The idea of text based variational autoencoders is shown in \cite{bowman-etal-2016-generating} with a combination of LSTM \cite{hochreiter1997long} and VAE. A more syntax informative text based VAE is used in \cite{zhang-etal-2019-syntax-infused} with two  distinct latent spaces. There are existing work on controllable text generation from latent space \cite{hu2018controlled}, \cite{zhao2017learning}. Prior to our work, there are several Transformer based approaches. In \cite{8852155}, authors use transformer based variational autoencoder with promising modifications. A conditional design of variational Transformer is proposed in \cite{ijcai2019-727} for story completion. 
Posterior collapse is common problem in text VAEs. This problem occurs when the decoder ignores the latent vector and generation is done by only decoder, like an auto-regressive language model. The study of preventing posterior collapse is discussed in \cite{park-lee-2021-finetuning}, \cite{Long2019PreventingPC} and \cite{li2020optimus}. We refer the reader to \cite{Lucas2019UnderstandingPC}, \cite{park-lee-2021-finetuning}, for a detailed study on the properties of posterior collapse and its relation to text based Variational Autoencoders.

\section{Proposed Method}
\subsection{Background}
An Autoencoder is a model for reconstructing the input. Reconstructing includes encoding the input $\mathbf{x}$ into a smaller space, which is called latent vector $z$, and decoding it to input space. The objective is to minimize the metric loss $d(\cdot)$ between input $\mathbf{x}$ and reconstructed variable $\hat{\mathbf{x}}$. When the objective is to represent the latent variable $z$ in terms of a known distribution, Autoencoders are not always working. If the distribution of $z$ is known, only the decoder of Autoencoder is enough to generate new samples by passing $z$ as input.

As a solution for this problem, Variational Autoencoder is proposed in \cite{kingma2014autoencoding}. It is used for learning latent space representation via Variational Bayes. Since we do not know the exact distribution of the input, our aim is to approximate the unknown distribution to a known distribution. This procedure is defined as Variational Bayes in literature. We thus choose a normal distribution $p(z) \sim \mathcal{N}(0, \mathbf{I})$ as a known distribution, and approximate distribution $p_\text{encoder}(\mathbf{z} \mid \mathbf{x}^{(i)})$ to this normal distribution at training time. Choosing simple and known distribution helps us to generate samples from latent space $\mathbf{z}$ by passing it to the probabilistic decoder $p_\text{decoder}$. Because, we learn to encode the input to normal distribution, then decode it to same space.

Thus, objective is to maximize the evidence lower bound (ELBO), which is defined as 
\begin{equation} \label{eq:1}
    \mathop{\mathbb{E}}[\log p_\text{decoder}(\mathbf{x}^{(i)} \mid \mathbf{z})] - \mathbb{KL}(p_\text{encoder}(\mathbf{z} \mid \mathbf{x}^{(i)}) \mid \mid p(\mathbf{z}))
\end{equation}
where $KL$ refers to Kullback–Leibler divergence, $\mathbf{x}$ is the input and $\mathbf{x}^{(i)}$ is the $i$th sample in minibatch $B$. $p_\text{encoder}(\cdot)$ and $p_\text{decoder}(\cdot)$ are simply feed forward layers in deep learning, for encoding and decoding.

\subsection{Modeling Class Conditional Variational Transformer}
According to the Equation (\ref{eq:1}); we desing the Transformer model, introduced in \cite{NIPS2017_3f5ee243}, by injecting conditional variational autoencoder between its encoder and decoder. To formalize the idea, given a sequence of $N$ tokens, $(t_1, t_2, ..., t_N)$, the encoder of the Transformer model is defined as $Encoder: t_i \in \mathbb{R}^{N} \rightarrow \mathbf{c_i} \in \mathbb{R}^{N \times D}$ where $D$ is the hidden dimension, and computes the contextual representation $\mathbf{c}_i$ of each token $t_i$:
\begin{equation}
    \mathbf{c}_1, \mathbf{c}_2, ..., \mathbf{c}_N = Encoder(t_1, t_2, ..., t_N)
\end{equation}
In the original Transformer model, contextual information from encoder flows through to the decoder's cross attention layer by mapping it to key and value. However, recent state-of-the-art language models \cite{devlin-etal-2019-bert}, \cite{liu2019roberta}, \cite{sanh2020distilbert} show that full sentence information can be obtained by only using contextual information from first token $t_1$ (correspondingly the \texttt{[CLS]} token). In the same way, we only use $\mathbf{c_1}$ token representation. 
The idea of conditional variational autoencoder starts from vector $\mathbf{c}_1$. First, the probabilistic encoder $p_{encoder}(\mathbf{z} \mid \mathbf{c}_1)$ of CVAE encodes sentence representation to a latent vector $\mathbf{z}$ with dimension of $L$. Instead of conditioning only latent vector $\mathbf{z}$, decoder is also conditioned to the class information of our sentence. This conditional objective should be done by interacting latent vector with conditional information. If we pass this class information to the decoder, the reconstruction is conditioned both on the latent vector and class labels. Prior work on conditional variational autoencoder sets an another condition vector \cite{https://doi.org/10.48550/arxiv.1703.00955} and replace entries with condition features \cite{Lim2018}. Instead of using two vectors, we replace first entry of latent vector with class integer $C$ to inject class information to decoding process. Denoting replaced vector $\mathbf{z}$ as $\mathbf{z}'$, decoder is reformulated as
\begin{equation}
    p_{decoder}(\mathbf{c}_1 \mid \mathbf{z}')
\end{equation}

Normally, the decoder of Transformer requires a whole sequence, however we use only vector $\mathbf{c}_1$ as input to VAE's encoder. After decoding in VAE, the reconstructed variable is copied $N$ times to create a sequence like the output of encoder of Transformer. This sequence of vectors is passed to the decoder of Transformer as key and value. 

Computationally, this formulation is more efficient than using whole information from encoder's output. It reduces the number of parameters of the model. We experiment that if we use whole output of the encoder, the decoder of Transformer tends to copy the sentences in training set. We thus believe that this methods acts a role as regularizer. Besides that, authors of \cite{https://doi.org/10.48550/arxiv.1911.03976} and \cite{park-lee-2021-finetuning} experimented that pooling the output of Transformer's encoder prevents posterior collapse. By following that, we use only $\mathbf{c}_1$ vector as a pooling strategy, to strengthen our model against posterior collapse.

The decoder part of our Transformer has same architecture with original Transformer. To make a formal definition between repeat vector $\mathbf{c}_i'$, the decoder computes logit vectors:
\begin{equation}
    \mathbf{l}_1, \mathbf{l}_2, ..., \mathbf{l}_N = Decoder(\mathbf{x}', \mathbf{c}_i')
\end{equation}
where $\mathbf{x}'$ is the shifted $\mathbf{x}$. These logits vectors are simply our final predictions $p$ in Figure \ref{fig:cvaet}. Each logit corresponds to a token from the vocabulary.

At inference time, conditional Transformer takes a random normal vector $\mathbf{z}$ with a class label $z_1 \leftarrow C$. This class label is chosen by desired class of the sentence to be generated. Vector is passed to decoder of VAE. At the same time, decoder of Transformer takes \texttt{[START]} token to decode it autoregressively.
\begin{table*}[t]
\caption{Results for IMDB and COLA on different models}
\resizebox{\linewidth}{!}{
\begin{tabular}{lccccccccc}
\toprule
\multirow{2}{*}{\textbf{Datasets}} & \multicolumn{2}{c}{\textbf{BERT}}                             & \multicolumn{2}{c}{\textbf{ALBERT}}                         & \multicolumn{2}{c}{\textbf{XLNet}}                           & \multicolumn{2}{c}{\textbf{DistilBERT}}                       \\ 
                                   & \multicolumn{1}{c}{\textit{Baseline}}   & \textit{Ours}       & \multicolumn{1}{c}{\textit{Baseline}}  & \textit{Ours}      & \multicolumn{1}{c}{\textit{Baseline}}   & \textit{Ours}      & \multicolumn{1}{c}{\textit{Baseline}}  & \textit{Ours}        \\ \midrule
\textbf{IMDB}                      & \multicolumn{1}{c}{79.96 $\pm$ 0.686} & 80.82 $\pm$ 0.379 & \multicolumn{1}{c}{78.74 $\pm$ 0.17} & 79.28 $\pm$ 0.23 & \multicolumn{1}{c}{-}                   & -                  & \multicolumn{1}{c}{79.24 $\pm$ 0.11} & 80.05 $\pm$ 0.066 \vspace{0.2cm} \\ 
\textbf{CoLA}                      & \multicolumn{1}{c}{70.31 $\pm$ 0.103}  & 70.73 $\pm$ 0.109  & \multicolumn{1}{c}{-}                  & -                  & \multicolumn{1}{c}{68.68 $\pm$ 0.438} & 68.9 $\pm$ 0.331 & \multicolumn{1}{c}{68.8 $\pm$ 0.273} & 69.61 $\pm$ 0.391  \\ 
\bottomrule
\end{tabular}
}
\label{tab:randn}
\end{table*}
\section{Experimental Setup}
We evaluate our model on different benchmarks with different pre-trained language models. 
\subsection{Benchmarks}
\begin{table}[b]
\caption{different number of sentences based on different train samples with Different Data Augmentation Methods}
\resizebox{\columnwidth}{!}{%
\begin{tabular}{lcccc}
\toprule
\textbf{Dataset} & \textbf{\# Train Samples} & \textbf{\# Test Samples} & \multicolumn{1}{l}{\textbf{\# Generated Sentences}}               \\ \midrule
\textbf{IMDB}             & 25000                     & 25000                    & 5000                                                               \\ 
\textbf{CoLA}             & 8551                      & 1043                     & 5000                                                               \\ 
\textbf{Rotten Tomatoes}  & 8530                      & 1066                     & \begin{tabular}[c]{@{}c@{}}10000 \\ (FWS, FWS + R, R)\end{tabular} \\ 
\textbf{TREC-10}           & 5452                      & 500                      & \begin{tabular}[c]{@{}c@{}}6000 \\ (FWS, FWS + R, R)\end{tabular}  \\ \bottomrule
\end{tabular}
}
\label{tab:data}
\end{table}

\subsubsection{CoLA}
The Corpus of Linguistic Acceptability (CoLA) is one of the presented benchmarks in General Language Understanding Evaluation (GLUE) \cite{wang-etal-2018-glue}. The task is to determine whether given sentence is a grammatically correct English sentence.
\subsubsection{TREC-10}
TREC-10 \cite{li-roth-2002-learning} is a collection of questions that are annotated with its question type. It is a multiclass classification task with class labels of abbreviation, human, entity, location, numeric, description.
\subsubsection{Rotten Tomatoes}
Rotten Tomatoes \cite{Pang+Lee:05a} is a movie review dataset which is annotated for binary classification on highly abstract sentiment representations. 
\subsubsection{IMDB}
IMDB dataset \cite{maas-EtAl:2011:ACL-HLT2011} is a large collection of movie review dataset. The task is to classify whether given sentence is positive review or not.
\subsection{Sentence Generation Techniques}
We follow different sentence generation techniques on different benchmarks for data augmentation. For example, we generate 5000 different sentences for CoLA dataset by using all training samples. On the other hand, since IMDB dataset contains 25000 training samples, we perform random sub-sampling on this dataset before training due to computation time characteristic of data augmentation techniques.

This random generation process is done by passing \texttt{[START]} token to the decoder, then decoding it in autoregressive fashion until seeing \texttt{[END]} token. We see that this generated sentences without any rule tends to begin with the most frequent starting words (for example "A", "The", "I") in trained dataset.

\begin{table*}[h!]
\caption{Results for Rotten Tomatoes and TREC-10 dataset on BERT model}
\resizebox{\linewidth}{!}{%
\begin{tabular}{lcccccccc}
\toprule
\textbf{Datasets}        & \textbf{Baseline}    & \textbf{Synonym Replacement} & \textbf{Contextual Replacement} & \textbf{Back-Translation} & $\mathbf{R}$            & $\mathbf{FWS + R}$     & $\mathbf{FWS}$                 \\ \midrule
\textbf{Rotten Tomatoes} & 74.27 $\pm$ 0.357 & 74.73 $\pm$ 0.622         & 75.77 $\pm$ 0.162            & 75.84 $\pm$ 0.649      & 75.58  $\pm$ 0.746 & 76.08 $\pm$ 0.425 & \textbf{76.3 $\pm$ 0.324} \vspace{0.2cm}\\ 
\textbf{TREC-10}          & 75.8 $\pm$ 0.113   & 78.1 $\pm$ 0.487           & 78.87 $\pm$ 0.523            & 79.2 $\pm$ 0.392       & 78.9 $\pm$ 0.408   & 79.73 $\pm$ 0.471 & \textbf{79.87 $\pm$ 0.57} \\ \bottomrule
\end{tabular}
}
\label{tab:fwp}
\end{table*}

To measure importance and contribution of first word, we choose a third sentence generation technique: pre-sampling first word of a generated sentence. To select less frequent words, we calculate frequency of each starting word and normalize it to get their density function. Then, we raised this distribution to the $3/4$rd power empirically. This allows us to generate sentences that are relatively distinct from characteristics of current sentences in dataset by passing \{\texttt{[START]}, $sample(p_{w_1}, ..., p_{w_S})$\} to the decoder at inference time.

However, since the aim is to generate a sentence that belongs to the class $C$, third technique is not sufficient for such datasets. For example, if we want to generate a sample belongs to class "human" in TREC-10 dataset, it is not quite right to start the sentence with "where". On account of this, for TREC-10, we sample the first word from distinct distributions related to desired class of sentence to be generated.

As shown in Table \ref{tab:data}, different number of sentences that are generated based on different train samples. On the other hand, three set of sentences are generated for Rotten Tomates and TREC-10. First Word Sampling ($FWS$) stands for first word of all generated sentences are sampled from described distribution. Random ($R$) stands for sentences are generated randomly, without any pre-determined set of rules. At last, First Word Sampling + Random ($FWS + R$) means half of the generated sentences are $FWS$, other half is $R$.

All generation techniques include class conditional latent vector.

\begin{algorithm}[t]
\caption{Class Conditional Sentence Generation}\label{alg:cap}
\hspace*{\algorithmicindent} \textbf{Input:} model, $C$, use\_fws\\
 \hspace*{\algorithmicindent} \textbf{Output:} a new sentence which belongs to class $C$
\begin{algorithmic}[1]
\State sample $z$ from $\mathcal{N}(0, \mathbf{I})$
\State $z[1]$ $\leftarrow C$
\State vae\_output $\leftarrow$ model.vae\_decoder($z$)
\If{use\_fws == True}
    \State $fw \leftarrow$ sample a first word from normalized frequencies
\Else
    \State $fw \leftarrow$ None
\EndIf
\State new\_sentence $\leftarrow$ model.transformer\_decoder(vae\_output, $fw$)
\State \Return new\_sentence
\end{algorithmic}
\end{algorithm}
\section{Results}
In this section, results are reported on 4 different benchmarks as stated in Section IV. For IMDB, TREC-10 and Rotten Tomatoes; results are reported on test set. For CoLA, we use dev set. To show the performance and contribution of our model, we finetune several pre-trained models (\cite{bert-base-uncased}, \cite{distilbert-base-uncased}, \cite{xlnet-base-cased}, \cite{albert-base-v1}) on original datasets and augmented datasets to compare. We run each experiment three times and calculate the margin of error with 95\% confidence level to provide statistically significant evidence. We generate an equal number of sentences for each class.

For IMDB and CoLA, we finetune BERT \cite{devlin-etal-2019-bert} and DistilBERT \cite{sanh2020distilbert}. Additionaly, we finetune ALBERT \cite{lan2020albert} for IMDB; XLNet \cite{NEURIPS2019_dc6a7e65} for CoLA.

Table \ref{tab:randn} shows accuracy and margin of error values on IMDB and CoLA dataset using different pre-trained models. For CoLA dataset, we train conditional variational Transformer and generate 5000 samples to expand the dataset. Although our method increases the performance compared to training with original dataset, this increase is very small. Since linguistic acceptability is affected negatively from very small changes, we believe that this is due to characteristics of the dataset.

On IMDB dataset, we train conditional variational Transformer with randomly chosen 7000 samples over 25000 samples. After that, 5000 samples are generated to expand the dataset. As shown in Table \ref{tab:randn}, this augmentation method has considerable performance increase compared to CoLA experiments.

We choose three different augmentation techniques to compare with our method on TREC-10 and Rotten Tomatoes. The first method is to replace a randomly chosen work with its synonym on WordNet \cite{10.1145/219717.219748}. The second one is to replace a randomly chosen word with most similar word using contextual word embeddings \cite{roberta-base}. The third method is back-translation. Sentences are translated into German and then back into the English \cite{facebook-wmt19-en-de}. 

For each method, the same numbers of sentences are generated as in Table \ref{tab:data}. Each sample is augmented in train dataset by iterating, if the size of augmentation is larger than number of samples in train set, we choose random samples to complete.

Table \ref{tab:fwp} shows accuracy and margin of error values on TREC-10 and Rotten Tomatoes datasets using only BERT model. For these two dataset, we prepare 3 different augmented datasets based on different generation techniques ($R$, $FWS + R$, $FWS$) as explained in Section IV. Experiments show that back-translation outperforms synonym and contextual word replacement. For both datasets, random generation ($R$) is below the other methods in terms of accuracy. 

There is considerable performance increase if we generate sequences with first word sampling ($FWS + R$ and $FWS$) compared with baseline and other augmentation methods.

%
\section{Discussions}
We propose a novel generative data augmentation method, which can generate new sentences from desired class label. Our proposed model has several advantages over other augmentation methods.

If a generative model for data augmentation generates sentences without any class condition, there can be a corruption between class distributions. Conditional variational Transformer keeps class information while encoding semantic and syntactic properties of language on both binary and multiclass classification datasets.

Generation of noised and corrupted sentences is inevitable outcome when it comes to sentence generation. We believe that this sentences plays a role as regularizer besides high quality generated sentences.

Conditional variational Transformer retains the advantages of generative language models. We can generate a sentence from given sequence with class conditionality. We calculate class based first word frequencies to generate new sentences over less frequent words. Our experiments shows this approach outperforms any random data generation procedure.

Experiments states that, our method performs well if the number of samples in dataset is limited. We double the number of samples in Rotten Tomatoes and TREC-10 datasets. This suggests that diversity can also be increased in small datasets. Examples of sentences produced are given in Table \ref{tab:exs} in appendix.

\section{Future Work}
Since we apply our method to classification tasks, there are several tasks in NLP. We believe that our method can be expanded to other tasks such as sequence labeling, textual entailment \footnote{Our models and promising future work implementations are publicly available at https://github.com/safakkbilici/Conditional-Variational-Transformer.}. 

On the other hand, we don't pre-train any models before training on presented benchmarks. It is possible to encode more linguistic information via unconditional latent space learning on big corpus. We plan to investigate the pre-training objective and performance on downstream tasks in future work.

\section{Acknowledgment}
This research is supported by TUBITAK, The Scientific and Technological Research Council of Turkey with project number 120E100. Also, we would like to thank to inzva community for providing us a valuable research environment and GPU support.
\bibliographystyle{IEEEtran}
\bibliography{IEEEabrv,mybib.bib}
\newpage
\clearpage
\begin{table*}[!t]
\caption{randomly chosen generated sentences for each dataset and class label}
\resizebox{\linewidth}{!}{%
\begin{tabular}{cl|c}
\toprule
\textbf{Datasets}        & \multicolumn{1}{c|}{$\mathbf{R}$}                                                                                                                                                                                                                                                                                                                                                                                                                                                                                                                                                                                                                                                                          & $\mathbf{FWS}$                                                                                                                                                                                                                                                                                                                                                                                                                                                                                                                                                                                                                                                                                         \\ \midrule \midrule
\textbf{IMDB}            & \begin{tabular}[c]{@{}l@{}}\textbf{\textit{Negative (0)}}:\\    This is a weak film I have rarely seen many of movies but \\    there was no plot to the same plot and as this film is simply awful.\\    The acting is poor the the camera work is just the nonexistent. \\    The acting is awful the cinematography is awful the only acceptable acting is pathetic.\\ \\ \textbf{\textit{Positive (1)}}:\\    This is a great movie I gave it a 10 out of 10 then that it is a little more realistic movie \\    than most of the actors are great actors and actresses like they added a shooting story. \\    The screenplay is very well done. The story in this movie is very well.\end{tabular} & \textbf{-}                                                                                                                                                                                                                                                                                                                                                                                                                                                                                                                                                                                                                                                                  \\ \hline 

\textbf{CoLA}            & \begin{tabular}[c]{@{}l@{}}\textbf{\textit{Unacceptable (0)}}:\\    The boys guardians employer we elected president.\\ \\ \textbf{\textit{Acceptable (1)}}:\\    The students demonstrated the technique this morning.\end{tabular}                                                                                                                                                                                                                                                                                                                                                                                                                                                                     & \textbf{-}                                                                                                                                                                                                                                                                                                                                                                                                                                                                                                                                                                                                                                                                                           \\ \hline
\textbf{Rotten Tomatoes} & \begin{tabular}[c]{@{}l@{}}\textbf{\textit{Negative (0)}}:\\    The movie is a negligible work of manipulation but its so mechanical.\\ \\ \textbf{\textit{Positive (1)}}:   A shimmeringly lovely comingofage portrait of family and finally roll \\    that could have viewers right out.\end{tabular}                                                                                                                                                                                                                                                                                                                                                                                                 & \multicolumn{1}{l}{\begin{tabular}[c]{@{}l@{}}\textbf{\textit{Negative (0)}}:\\    Ordinary melodrama with weak dialogue and exactly as the direction of its lead performances.\\ \\ \textbf{\textit{Positive (1)}}:\\    Showtime is a film thats destined to win a documentary that is a certain degree to raise.\end{tabular}}                                                                                                                                                                                                                                                                                                                                                                   \\ \hline
\textbf{TREC-10}          & \begin{tabular}[c]{@{}l@{}}\textbf{\textit{Description (0)}}:\\    What is the difference between pop music to drink?\\ \\ \textbf{\textit{Entity (1)}}:\\    What is the correct way to use with James home with letter?\\ \\ \textbf{\textit{Abbreviation (2)}}:\\    What is the abbreviation of the computer at General Motors?\\ \\ \textbf{\textit{Human (3)}}:\\    Who is the first Taiwanese President to be popular sleep?\\ \\ \textbf{\textit{Numeric (4)}}: \\    What is the temperature for having a typist to a beach?\\ \\ \textbf{\textit{Location (5)}}:\\    What is the name of the Wilkes plantation in the Ewoks live on\end{tabular}                                             & \multicolumn{1}{l}{\begin{tabular}[c]{@{}l@{}}\textbf{\textit{Description (0)}}:\\    How can I find out how much much do food if I know their questions?\\ \\ \textbf{\textit{Entity (1)}}:\\    What is the best way to remove?\\ \\ \textbf{\textit{Abbreviation (2)}}:\\    When reading classified ads what does EENTY other stand for?\\ \\ \textbf{\textit{Human (3)}}:\\    What was the name of the American who was known for a senator in The Roman Black swearing and 50s when?\\ \\ \textbf{\textit{Numeric (4)}}: \\    When did the art live?\\ \\ \textbf{\textit{Location (5)}}:\\    Where can I find a website about climbs of Mount Everest and sailors monument?\end{tabular}} \\ \hline
\end{tabular}
}
\label{tab:exs}
\end{table*}

\begin{table}[b]
\caption{Variable Hyperparameters for Each Dataset. MSL stands for Maximum Sequence Length.}
\begin{tabular}{lcccc}
\toprule
\textbf{Dataset}         & \textbf{MSL} & \textbf{Batch Size} & \textbf{Epochs} & \textbf{Learning Rate} \\ \midrule
\textbf{IMDB}            & 100          & 32                  & 50              & 1E-4                  \\ 
\textbf{CoLA}            & 70           & 64                  & 95              & 2E-4                   \\ 
\textbf{Rotten Tomatoes} & 100          & 64                  & 95              & 2E-4                  \\ 
\textbf{TREC-10}          & 50           & 64                  & 70              & 2E-4                  \\ \bottomrule
\end{tabular}
\label{tab:hy}
\end{table}

\appendix


\subsection{Generated Sentence Examples}
In this section, we show one randomly chosen generated sentence for each dataset and their class labels. For Rotten Tomatoes and TREC-10, $FWS$ samples are shown as well. In Table \ref{tab:exs}, it is clear that our model can distinguish negative and positive sentiment for IMDB and Rotten Tomatoes datasets.

If we examine the characteristics of movie review datasets, it has high probability to start a sentence with "this movie" phrase. With introducing First Word Sampling approach $PWS$, diverse reviews can be generated.

On the other hand, our model captures the logical representation of question types. Even though the TREC-10 dataset has limited number of train samples, our model is capable to generate sentences which are not seen in train set.

\subsection{Hyperparameters \& Training Details}
We use same hyperparameters for each dataset. Dimensions of keys, queries and values are set to 64.  Dimension of word vectors $d_{model}$ is set to 256. We use 3 layers for encoder and decoder. For each multi-head attention layer, number of heads is set to 8. We apply dropout with rate of 0.1. We don't use any learning rate scheduler. Adam optimizer \cite{kingma2017adam} is used with $\beta_1=0.9$, $\beta_2=0.999$. At training time, we apply label smoothing \cite{7780677} with $\epsilon_{ls} = 0.1$. Other hyperparameters are given in Table \ref{tab:hy}.

To prevent KL cost annealing, we follow the same technique in \cite{bowman-etal-2016-generating}. We introduce a variable weight $w_0$ for KL term in the loss, then the weight is increased gradually from 0 to 1 at training. We update value of the weight with a simple logistic function
\begin{equation}
    w_0 = \frac{1}{1 + \exp(-k \cdot (t - x_0))}
\end{equation}
where $k=0.0025$, $x_0 = 2500$ and $t$ is the current step.
\end{document}